\documentclass{article}

\PassOptionsToPackage{numbers}{natbib}
\usepackage[dblblindworkshop, final]{neurips_2026}
\workshoptitle{Personalized, Aligned, Long-Term Memory for AI Systems}

\usepackage[utf8]{inputenc}
\usepackage[T1]{fontenc}
\usepackage{hyperref}
\usepackage{url}
\usepackage{booktabs}
\usepackage{amsfonts}
\usepackage{nicefrac}
\usepackage{microtype}
\usepackage{xcolor}

\usepackage{amsmath,amssymb}
\usepackage{graphicx}
\usepackage{multirow}
\usepackage{tikz}
\usetikzlibrary{shapes,arrows.meta,positioning,fit,calc}
\usepackage{enumitem}
\usepackage{caption}
\usepackage{subcaption}
\usepackage{algorithm}
\usepackage{algpseudocode}
\usepackage{tcolorbox}
\usepackage{tabularx}
\usepackage{float}
\usepackage{longtable}
\usepackage{makecell}
\usepackage{pifont}
\newcommand{\cmark}{\ding{51}}

\definecolor{dcblue}{HTML}{2563EB}
\definecolor{dcgreen}{HTML}{059669}
\definecolor{dcred}{HTML}{DC2626}
\definecolor{dcorange}{HTML}{D97706}
\definecolor{dcgray}{HTML}{6B7280}

\usepackage{xspace}

\newcommand{\forgedc}{\texttt{forge\_dc}\xspace}
\newcommand{\luciddc}{\texttt{lucid\_dc}\xspace}
\newcommand{\hybridraw}{\texttt{hybrid\_raw}\xspace}
\newcommand{\docchunk}{\texttt{doc\_chunk}\xspace}
\newcommand{\docnaive}{\texttt{doc\_naive}\xspace}

\newcommand{\Rule}{\textsc{rule}\xspace}
\newcommand{\Format}{\textsc{format}\xspace}
\newcommand{\Content}{\textsc{content}\xspace}

\title{Decision Context: Persistent Behavioral Learning for AI Agents}

\author{%
  Nima Vahdat\thanks{Currently at PwC. \texttt{nima.vahdat@pwc.com}} \\
  Georgian AI Lab \\
  \texttt{nima.vahdat@georgian.io}
  \And
  Pashootan Vaezipoor \\
  Georgian AI Lab \\
  \texttt{pashootan@georgian.io}
}

\begin{document}

\maketitle

\begin{abstract}

LLM agents do not learn from corrections across sessions. A user fixes
the same mistake at the start of every conversation. Existing agent
memory systems handle facts (preferences, entities, conversational
history~\cite{packer2023memgpt, chhikara2025mem0, rasmussen2025zep})
but not procedural knowledge~\cite{ai2025memorybench}.

We present Forge Decision Context (\forgedc), a
system that captures behavioral signals from human corrections and from
self-assessed tool-call failures, compiles them into imperative
\Rule directives, and injects them at the decision points where
they apply. We evaluate \forgedc alongside four predecessor
approaches on \textbf{DC-Bench}, a new benchmark of 71 scenarios across
9 behavioral dimensions, built from real executed agent traces in a
deterministic mocked-tool environment. \forgedc
reaches 95\% compliance, up from 55\% with naive retrieval, while
injecting roughly $10\times$ fewer tokens per query.

Across the five-stage progression, the strongest systems represent
behavioral memory as explicit directives and deliver it at the decision
points where it applies. The results identify representation,
activation, and placement as central design choices for persistent
behavioral learning.

\end{abstract}

\section{Introduction}
\label{sec:introduction}

LLM agents are increasingly deployed beyond question-answering: writing
code, managing infrastructure, orchestrating multi-step workflows. Yet they
are \emph{memoryless learners}: every session begins from the same blank
state. Every correction must be repeated. Every hard-won behavioral
insight is lost.

Consider a security analyst correcting an agent over many sessions:
``\textit{always check Identity and Access Management (IAM) policies before network configurations},'' ``\textit{use
\texttt{--recursive} when scanning S3 buckets},'' ``\textit{that API returns
paginated results, so loop}.'' Within each session the agent improves as
corrections accumulate in context. But the next day every correction must
be re-issued, in the same order, with the agent making the same mistakes
with the same confidence.

The problem has two dimensions. First is the loss of \textbf{human
corrections}: explicit feedback transferring domain expertise about
metrics, parameters, and strategies. Second is the loss of
\textbf{self-generated patterns}: signals from the agent's own execution
such as tool retries that eventually succeed, errors that force strategy
changes, and outputs the user silently regenerates. Existing prompt-engineering
practice is fundamentally lossy (most corrections never propagate to
system prompts), situational~(most judgments are conditional), and
ignores self-generated signals entirely. What is needed is a system that
captures situational judgments as they occur and surfaces the right
judgment at the right moment.

The agent memory landscape has grown rapidly~\cite{hu2025memory,
du2026memory}. Many systems emphasize facts, such as a meeting on
Tuesday, a Python version, or a preference for dark mode. Other systems
extend memory beyond factual recall by learning user preferences from
edits~\cite{gao2024prelude} and reusable procedures from interaction
traces~\cite{chen2024automanual,zhou2026mementoskills,liu2026skillrevise}.
We study how a tool-using
agent can turn explicit corrections and implicit execution signals into
actor-scoped guidance, determine when that guidance applies, and deliver
it at the decision point where it can change behavior.
Storing the correction ``\textit{use QRR for enterprise revenue}'' is
trivial; knowing \emph{when}, \emph{how}, and \emph{whether} to apply it
is the hard problem.

A separate line of work learns reusable procedures, workflows, or hints from past trajectories, scored by task success~\cite{wang2024awm,
fang2025memp, zhang2025ace, nekoei2025jef}. These methods assume a reward: something outside the agent says which trajectory was better. Deployment provides no such signal. What the agent has is the session itself: an explicit human correction, or an implicit cue in its own execution, such as a tool call that fails and then succeeds on retry, or an output the user silently regenerates.

This paper makes three contributions:
\begin{enumerate}
\item \textbf{DC-Bench.} A benchmark of 71 scenarios across 9 behavioral dimensions and two agent domains, built from real traces captured by running agents in a deterministic mocked-tool environment. Each scenario seeds corrections on one subject and tests on another, so it measures whether a correction is \emph{applied}, not whether it is recalled.
\item \textbf{Forge Decision Context (\forgedc).} A system that learns from human corrections and self-assessed tool-call failures without a reward signal, compiles both into scoped, actor-attributed \Rule directives, and injects them at the decision points where they apply. It reaches 95\% compliance on DC-Bench, against 12--15\% for factual-memory systems and 34--49\% for procedural-memory baselines.
\item \textbf{Empirical findings from a five-stage system progression.} The progression reveals an advantage for directive-form memory, a retrieval gap between formatting and content rules, the failure of implicit signals without explicit extraction, and sensitivity to the injection channel (Section~\ref{sec:findings}).
\end{enumerate}

The implementation and evaluation artifacts are available at
\url{https://github.com/NimaVahdat/decision-context-memory} and
\url{https://github.com/NimaVahdat/decision-context-benchmark}.

\section{Related Work}
\label{sec:related-work}

\paragraph{Agent memory systems.}

Existing systems target factual memory in three families.
\emph{Graph-structured} systems (Zep/Graphiti~\cite{rasmussen2025zep},
GraphMem~\cite{yang2026graph}) use typed graphs with traversal-based
retrieval. \emph{LLM-as-controller} systems
(MemGPT/Letta~\cite{packer2023memgpt}, MemOS~\cite{memos2025}) page
tiered memory in and out via LLM tool calls.
\emph{Extraction-pipeline} systems (Mem0~\cite{chhikara2025mem0},
Hindsight~\cite{hindsight2025}) extract structured records offline with
an LLM, then retrieve via non-LLM hybrid search. These systems primarily
optimize factual or conversational recall; procedural memory remains
comparatively underexplored~\cite{du2026memory, logan2026continuum}.
DC is complementary: an agent may use Mem0 for facts and DC for
behavior simultaneously.

\paragraph{Procedural and experiential memory.}

A recent line distills reusable knowledge from past trajectories:
AWM~\cite{wang2024awm} induces workflows, Memp~\cite{fang2025memp}
builds step-level procedures, ACE~\cite{zhang2025ace} curates a context
playbook, and JEF-Hinter~\cite{nekoei2025jef} extracts offline hints.
AutoManual~\cite{chen2024automanual} constructs and maintains an online
rule manual, while skill-centric systems persist and revise reusable
procedures: Memento-Skills~\cite{zhou2026mementoskills} grows a routed
skill library, and SkillRevise~\cite{liu2026skillrevise} diagnoses and
repairs skills through re-execution. These systems primarily learn
procedures that expand or repair how an agent completes tasks. Decision
Context instead learns scoped behavioral guidance that changes when and
how an agent uses capabilities it already has. It converts explicit
corrections and patterns from tool-use traces into actor- and
condition-scoped directives delivered at relevant decision points.
Accordingly, skill-learning systems are generally evaluated by task
success, whereas DC-Bench measures conditional compliance,
generalization, scope precision, and non-transfer. We adapt AWM and Memp
as baselines because their trace-to-procedure pipelines most directly
match the behavioral-memory setting.

\paragraph{Learning preferences from edits.}

PRELUDE/CIPHER~\cite{gao2024prelude} infers context-dependent latent
preferences from users' direct edits and retrieves and aggregates those
preferences to guide future generation. It is the closest prior work on
reusing naturally occurring user revisions. Our setting broadens the
feedback beyond edited text to explicit conversational corrections and
tool-execution patterns, and focuses on multi-step tool agents where a
rule may need to activate between calls, be scoped to an actor, or be
suppressed to avoid negative transfer. Conversely, PRELUDE directly
optimizes edit cost in writing tasks, which DC-Bench does not measure.

\paragraph{Memory benchmarks.}

LongMemEval~\cite{wu2024longmemeval}, MABench~\cite{hu2026mabench},
MemoryBench~\cite{ai2025memorybench}, MemBench~\cite{membench2025},
MemoryArena~\cite{memoryarena2026}, and LOCOMO~\cite{locomo2024} converge
on four findings: (i)~single-fact retrieval is largely solved; (ii)~cross-session
reasoning remains difficult; (iii)~conflict resolution is weak across all
systems; (iv)~procedural memory is unsolved.
MemoryBench~\cite{ai2025memorybench} explicitly identifies the procedural
gap that motivates this work. DC-Bench differs by evaluating behavioral
\emph{compliance} (does the agent apply the correction to an unseen
subject?), using real executed agent traces with controlled tool outputs, and covering 9 behavioral
dimensions including self-assessment and multi-actor conflict. The
benchmark is intended to complement, not replace, task-success and
preference-learning evaluations.

\paragraph{Learning from feedback and RAG.}

RLHF~\cite{ouyang2022training} adjusts model weights at the population
level and is unsuitable for per-user behavioral adaptation.
Constitutional AI~\cite{bai2022constitutional} encodes manually
authored, universal principles; DC automates and personalizes the same
idea. In-context learning~\cite{brown2020language} requires manual
example selection and does not persist across sessions. Continual
learning~\cite{logan2026continuum} faces catastrophic forgetting and
training cost. Standard RAG~\cite{lewis2020rag} retrieves information
once at query start; DC retrieves behavioral guidance progressively at
the decision points where it applies.

\section{The Dimensions of Behavioral Memory}
\label{sec:problem-definition}
\label{sec:formal-problem}

Behavioral memory must satisfy several dimensions that factual memory
does not, and prior approaches address only some of them. The two
signal sources (human corrections and self-generated trace patterns)
must be turned into reusable guidance and surfaced when relevant. Six
challenges define these dimensions: extracting signals from noise,
generalizing without losing fidelity, retrieving at scale, surfacing at
the decision point, updating knowledge as corrections supersede one
another, and not regressing behavior the agent already gets right.

\paragraph{Extracting signals from noise.}

Behavioral corrections arrive as natural language embedded in
conversation. ``Oh, and include the USD symbol next to monetary amounts''
is a correction, but it looks like a casual remark. Trace patterns
present a different challenge: not every tool error is a behavioral
failure (rate limits, empty results, and sequential narrowing all
produce errors that are not signals).

\paragraph{Generalizing without losing fidelity.}

A correction observed against one subject must transfer to others
without dropping qualifiers. Three properties make this hard:
(i)~corrections \emph{chain}, so the same behavior is corrected
repeatedly and only the final state matters; (ii)~they are
\emph{conditional}, with qualifiers (``for enterprise clients'') that
must be preserved; (iii)~the generalization scope is ambiguous, since
a failure on one subject may or may not generalize to others.

\paragraph{Retrieval at scale.}

Behavioral guidance has narrow windows of relevance: the same tool with
a different source, the same metric for a different segment. Scale
stresses precision: the system must work with 5 corrections and with
500. Different categories also retrieve differently. Content-related
guidance has semantic overlap with the query; formatting guidance
(``include USD next to amounts'') does not, since ``USD formatting'' has
no embedding similarity to ``Beta Inc.''

\paragraph{Surfacing at the right moment.}

Unlike facts, behavioral guidance must appear at the decision point
where it is actionable. Guidance about S3 scanning is irrelevant during
planning and critical during execution. Premature injection wastes
context, late injection misses the decision point, and content placed
in the middle of long contexts attenuates~\cite{liu2024lost}.

\paragraph{Updating and resolving conflicts.}

Corrections accumulate, change, and conflict. A later correction
supersedes an earlier one, and different actors correct the same
behavior differently (one says ``QRR,'' another says ``ARR''). The
system must track provenance (\emph{who} said \emph{what}, \emph{when})
and keep the knowledge current both at maintenance time (deduplication,
supersession) and at injection time (selecting the right actor's
preferences).

\paragraph{Not regressing prior behavior.}

Injected guidance must not degrade outputs the agent already produces
correctly. A rule that fires on a case it does not apply to, or that
contradicts behavior already in place, can make the agent worse than no
memory at all, so the system needs a default of staying silent when
nothing needs to change.

\section{System Design}
\label{sec:approach}
\label{sec:approaches}

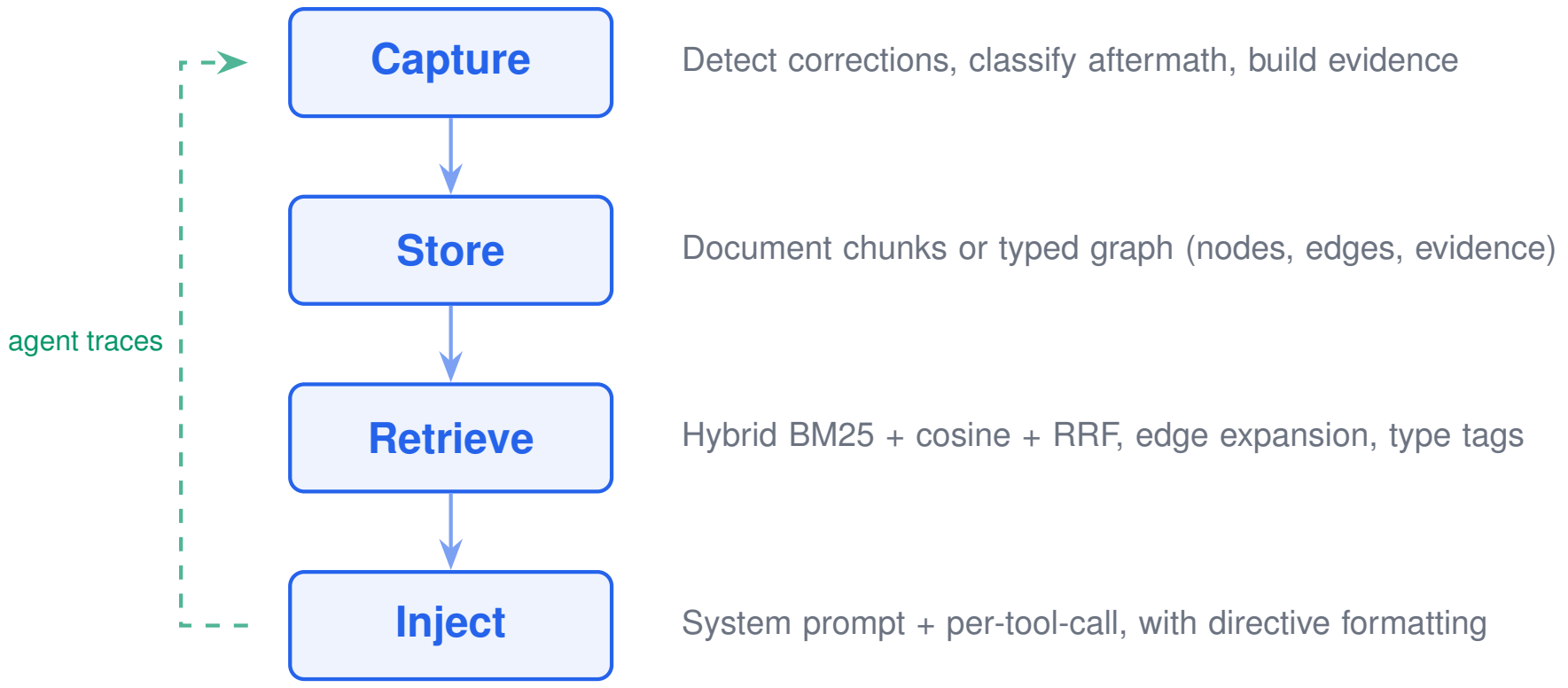
\begin{figure}[t]
\centering
\begin{tikzpicture}[
    box/.style={
        draw=dcblue, thick, rounded corners=3pt,
        minimum width=2.4cm, minimum height=0.8cm,
        fill=dcblue!8, font=\sffamily\small\bfseries, text=dcblue
    },
    desc/.style={font=\sffamily\scriptsize, text=dcgray, anchor=west, text width=8cm},
    arr/.style={-Stealth, thick, dcblue!60}
]
\node[box] (C) at (0, 0)    {Capture};
\node[box] (S) at (0, -1.4) {Store};
\node[box] (R) at (0, -2.8) {Retrieve};
\node[box] (I) at (0, -4.2) {Inject};
\node[desc] at (1.6, 0)    {Detect corrections, classify aftermath, build evidence};
\node[desc] at (1.6, -1.4) {Document chunks or typed graph (nodes, edges, evidence)};
\node[desc] at (1.6, -2.8) {Hybrid BM25 + cosine + RRF, edge expansion, type tags};
\node[desc] at (1.6, -4.2) {System prompt + per-tool-call, with directive formatting};
\draw[arr] (C) -- (S);
\draw[arr] (S) -- (R);
\draw[arr] (R) -- (I);
\draw[arr, dcgreen!70, dashed]
    ([xshift=-0.3cm]I.west) -- ++(-0.5,0) |- ([xshift=-0.3cm]C.west)
    node[font=\sffamily\tiny, text=dcgreen, pos=0.25, left] {agent traces};
\end{tikzpicture}
\caption{Four-layer Decision Context architecture. The dashed arrow is the
feedback loop: traces from injected sessions become input to capture.}
\label{fig:architecture}
\end{figure}

The six challenges in Section~\ref{sec:formal-problem} map to a four-layer
pipeline (Figure~\ref{fig:architecture}): \textbf{Capture} extracts
behavioral signals from agent traces; \textbf{Store} persists them in
queryable form; \textbf{Retrieve} finds relevant rules for a new query;
\textbf{Inject} delivers them at the right moment. Five approaches
implement this pipeline with progressively more sophistication
(Table~\ref{tab:approaches-overview}). \forgedc, the
strongest approach and the main technical contribution, combines all of
them.

\paragraph{Shared infrastructure.}

All five approaches share an embedding pipeline
(\texttt{text-embedding-3-small}) and hybrid search via Reciprocal Rank
Fusion~\cite{cormack2009rrf}
($\mathrm{RRF}(n) = 1/(k + r_{\mathrm{emb}}) + 1/(k + r_{\mathrm{BM25}})$,
$k=60$). Four of the five (all except \docnaive) additionally
implement \emph{aftermath classification}: deterministic keyword
detection (``wrong,'' ``actually,'' ``instead'') with LLM fallback that
decides whether each subsequent message is a correction, an acceptance,
or a new request. These chain boundaries matter for grouping related
corrections.

\paragraph{Two storage models.}

The first two approaches (\docnaive, \docchunk) use a
flat \texttt{Documents}~$\to$~\texttt{Chunks} schema. The remaining three
(\hybridraw, \luciddc, \forgedc) use a
typed graph: \texttt{DecisionNode}s carry both a \emph{node embedding}
(local content) and an \emph{accumulated embedding} (concatenated
content up to that node), enabling stage-to-stage matching. Eight base
edge types (\texttt{MADE}, \texttt{ABOUT}, \texttt{APPLIED},
\texttt{LED\_TO}, \texttt{PRODUCED}, \texttt{INFORMED\_BY},
\texttt{HAS\_SUMMARY}, \texttt{FOLLOWED\_BY}) connect actors, subjects,
nodes, and evidence; \forgedc adds a ninth (\texttt{FORGED\_FROM})
linking each compiled rule to the trigger node that should surface it.

\begin{table}[htbp]
\centering
\small
\caption{The five Decision Context approaches. Compliance scores are
averages across the 71 DC-Bench scenarios (Section~\ref{sec:experiments}).}
\label{tab:approaches-overview}
\begin{tabular}{@{}lccccc@{}}
\toprule
\textbf{Approach} & \textbf{Storage} & \textbf{Self-Assess} & \textbf{Compilation} & \textbf{Per-Tool} & \textbf{Avg.} \\
\midrule
\docnaive  & Chunk    & ---      & ---      & ---          & 55\% \\
\docchunk  & Annot.\ chunk & ---      & ---      & ---          & 63\% \\
\hybridraw & Graph    & ---      & ---      & Evidence     & 70\% \\
\luciddc   & Graph    & Sonnet 4 & ---      & Evidence     & 82\% \\
\forgedc   & Graph    & Sonnet 4 & Sonnet 4 & Re-retrieval & \textbf{95\%} \\
\bottomrule
\end{tabular}
\end{table}

\subsection{Four predecessor approaches}

\paragraph{\docnaive (55\%): naive RAG.}

The simplest pipeline: each trace is split at tool-call boundaries and
chunked; embed, retrieve the top-$k$ via the shared hybrid search
(BM25 + cosine + RRF), and inject as ``relevant past context'' in the
system prompt. No
annotation, no aftermath classification, no edges, no per-tool
injection. This is what happens when a team dumps logs into a vector
database. Corrections get fragmented across chunks, are mixed with
conversational filler, and are injected as raw text the agent must
interpret. The 55\% baseline succeeds only on simple, recent corrections
that happen to land in a retrievable chunk.

\paragraph{\docchunk (63\%): annotated chunks.}

The motivation here is to improve retrieval by adding annotations, and
see whether the same annotations can shrink the injected context.
\docchunk adds deterministic per-chunk annotations and
aftermath classification on top of \docnaive, and prepends a
global readout of per-action correction rates to the system prompt
alongside top precedents. Storage stays flat (no graph edges).
Annotation makes BM25 hits sharper and lets injection drop
conversational filler, raising compliance by 8 points (55\%→63\%); but
without edges the system still cannot follow correction chains, traverse
actor-scoped history, or use accumulated embeddings.

\paragraph{\hybridraw (70\%): graph storage.}

Replaces chunks with the typed-graph storage model: each session becomes
a sequence of \texttt{DecisionNode}s with eight edge types and both node
and accumulated embeddings. The shared hybrid search now operates over
nodes, with the embedding component combining node and accumulated
similarity ($0.7\cdot\text{node} + 0.3\cdot\text{accum}$). After RRF
fusion, multiplicative boosts re-rank candidates by matching subjects,
actors, and corrected documents. Injection occurs at two points: a
system-prompt global readout plus top precedents, and a separate
per-tool-call lookup that matches the current tool result against stored
evidence via \texttt{INFORMED\_BY} edges. Structure alone fixes retrieval (chains, conditional matching,
long-conversation and scale scenarios), but \hybridraw only
captures explicit human corrections. Implicit signals from the agent's
own execution, tool retries, recurring failures, and suboptimal
sequences, are never extracted. This sets up the next approach.

\paragraph{\luciddc (82\%): autonomous self-assessment.}

Asks whether the agent can learn from its \emph{own} traces rather than
human corrections. After base capture, a separate Sonnet~4 analyzer reads
\emph{tool-only} data (no conversational context) and compiles recurring
failures, retry-then-succeed sequences, and suboptimal strategies into
behavioral directives. Tool-only input is deliberate: when the analyzer
sees full traces, it produces rules about everything; restricting input
to tool calls keeps it focused on observable patterns. Compiled rules
are injected unconditionally under a \texttt{MANDATORY TOOL-CALLING
RULES} header, alongside the \hybridraw retrieval. Only the
self-assessment directives were followed reliably; raw human corrections
were often ignored. This observation motivated \forgedc's correction
compiler and the ``Only Rules Rule'' mnemonic
(Section~\ref{sec:findings}).

\subsection{\texorpdfstring{\forgedc}{forge\_dc}: correction compilation with progressive injection}

\forgedc combines both behavioral signal sources through a
compilation pipeline that transforms raw corrections and trace patterns
into structured, conflict-resolved, progressively injectable
\Rule directives.

\paragraph{Pipeline.}

On top of the \luciddc capture and self-assessment:
(1)~\textbf{Correction chains} are discovered by grouping per
\texttt{conversation\_id} and \texttt{actor\_id}, then splitting at
\emph{accepted} aftermath boundaries (per-actor grouping prevents
cross-preference merging).
(2)~\textbf{Compilation} (Sonnet~4) turns each chain, and the
self-assessment patterns, into structured output: \Rule
(imperative directive), \textsc{type} (\Format or
\Content), \textsc{trigger} (precondition node),
\textsc{drop}/\textsc{skip} (noise filters), and \textsc{group} markers
that cluster related rules.
(3)~Each rule is wired to its trigger node via a \texttt{FORGED\_FROM}
edge.
(4)~\textbf{Rulebook maintenance} deduplicates, merges agreements,
supersedes earlier rules from the same actor, and flags cross-actor
conflicts for runtime resolution.
(5)~At query time, \textbf{progressive injection} delivers rules across
two stages.

\paragraph{Precondition triggers.}

For each rule the compiler answers: \emph{at which point in the trace,
if the agent had known this rule, would the mistake have been
prevented?} For tool-specific rules this is the node \emph{before} the
failing call (so the rule fires in time to prevent the mistake). The
trigger anchors the \texttt{FORGED\_FROM} edge; at query time, matching
the agent's current context against trigger embeddings surfaces the
linked rule.

\paragraph{Progressive injection.}

Rules arrive at the model through two channels. \textbf{Stage~1}
(before the agent starts) injects unconditional rules (\Format
type, plus any rule whose trigger matches the initial query) through
the \emph{user channel} as a fabricated user/assistant dialogue pair
prepended to the conversation. \textbf{Stage~2+} (between tool calls)
re-retrieves using the agent's accumulated context (query plus all
prior tool calls and results) against trigger accumulated embeddings,
and adds each newly triggered rule to the \emph{system prompt} for the
next model call. The two channels are intentionally different
(Section~\ref{sec:channel-finding}). This stage-to-stage matching
surfaces data-conditional rules at the moment the triggering data
appears, and re-injecting at the decision point mitigates the ``lost
in the middle'' problem~\cite{liu2024lost}.

\paragraph{Conflict resolution.}

When retrieval surfaces rules from the same conflict group, resolution
is three-tiered: (i)~prefer the querying actor's rule if they are a
party to the conflict; (ii)~otherwise pick the majority rule by
supporting corrections; (iii)~if no majority, abstain. All rules are
stored, and selection happens at injection time per actor, allowing the
same rule set to serve different actors differently.

\section{DC-Bench}
\label{sec:benchmark}

DC-Bench evaluates \emph{behavioral} recall: can the system learn from
corrections and self-assessment, then \emph{apply} that learning to a
novel subject? It comprises 71 scenarios across 9 behavioral dimensions
(Behavioral Change, Conditional Activation, Exception Handling,
Generalization Boundary, Implicit Learning, Persistence, Recall, Scope
Precision, Temporal Coherence; per-scenario scores in
Appendix~\ref{app:per-scenario}), built on four principles. Each dimension
contains scenarios from both a financial analyst skill and a customer
support skill, testing whether behavioral learning transfers across
agent domains.

\paragraph{Real agent traces with controlled tools.}
Every seed is a real trace captured by running the agent end to end. The
trace records the model's actual messages, reasoning trajectory, tool
calls, tool results, and subsequent author-provided behavioral
corrections. The agent therefore determines which tools to call, in what
order, and how to react to their results. Only the external environment
is controlled: tools are mocks whose outputs are predetermined for each
scenario, ensuring that every approach encounters the same evidence
while preserving natural action sequences and trace structure.

\paragraph{Seed-then-test on unseen subjects.}
Each scenario seeds corrections for one company (Acme) and queries about
a \emph{different} company (Beta), forcing behavioral generalization.

\paragraph{Decoupled evaluation.}
Each approach captures with its own pipeline, but the test phase uses an
identical agent (same skill prompts, tools, LLM config). The only
variable at test time is the injected context, so score differences
reflect retrieval and injection quality.

\paragraph{Scenario isolation.}
Every scenario starts with a fresh memory store. Memories do not carry
across scenarios, preventing contamination from unrelated corrections
and ensuring that each score reflects only the seed trace specified for
that scenario. The five repeated runs share the scenario specification
but do not inherit memory from another scenario.

\paragraph{Dual scoring and compliance.}
A GPT-4 judge evaluates each output against a 3--8-item rubric (positive
or negative mode, weighted 1--3). The satisfied rubric weight is
normalized by the applicable total weight to produce a per-run
compliance score between 0 and 100. We average the five runs for each
scenario--approach pair, then macro-average scenario scores within each
dimension and across all 71 scenarios. Programmatic checks verify
deterministic constraints (\texttt{tool\_call\_max},
\texttt{source\_not\_called}, \texttt{output\_contains},
\texttt{tool\_param\_check}); programmatic failure forces a~0. Each
scenario--approach pair is evaluated over 5 runs.

\section{Results}
\label{sec:experiments}

The main results use Opus~4.7 as the agent backbone, GPT-4 as the judge,
and \texttt{text-embedding-3-small} for embeddings, with 5 runs per
scenario--approach pair (mean reported). \texttt{mem0} and \texttt{zep}
are factual-memory baselines; \texttt{awm}~\cite{wang2024awm} and
\texttt{memp}~\cite{fang2025memp} are procedural-memory baselines, given
the identical seed traces and injection channel so any difference
reflects what each system learns.
Table~\ref{tab:category-results} reports mean compliance by behavioral
dimension.

\begin{table}[htbp]
\centering
\small
\setlength{\tabcolsep}{3pt}
\caption{Mean compliance (\%) by behavioral dimension across 71 scenarios.
\texttt{ctrl} = no injection. \texttt{mem0}/\texttt{zep} are factual
memory baselines; \texttt{awm}/\texttt{memp} are procedural-memory
baselines. Best per row \textbf{bolded}.
$\Delta$ = \texttt{forge} $-$ \texttt{naive}.}
\label{tab:category-results}
\begin{tabular}{lcccccccccc r}
\toprule
\textbf{Dimension} & \texttt{ctrl} & \texttt{mem0} & \texttt{zep} & \texttt{awm} & \texttt{memp} & \texttt{naive} & \texttt{chunk} & \texttt{hybrid} & \texttt{lucid} & \texttt{forge} & $\Delta$ \\
\midrule
Behav.\ Change (7)     & 14 & 14 & 14 & 43 & 71 & 71 & 71 & \textbf{86} & \textbf{86} & \textbf{86} & +15 \\
Cond.\ Activation (12) &  8 &  0 & 12 & 49 & 40 & 53 & 76 & 80 & 84 & \textbf{95}  & +42 \\
Exception (3)          & 67 & 67 & 67 & 67 & \textbf{100} & 67 & \textbf{100} & 87 & 87 & \textbf{100} & +33 \\
Generalize (3)         & 67 & 33 & 33 & 33 & \textbf{100} & 67 & 67 & 67 & 67 & \textbf{100} & +33 \\
Implicit Learn.\ (10)  &  0 &  0 &  0 &  0 & 10 & 10 & 20 & \emph{0} & 78 & \textbf{90} & +80 \\
Persistence (10)       &  8 &  2 &  5 & 41 & 21 & 21 & 37 & 82 & 83 & \textbf{100} & +79 \\
Recall (9)             &  0 &  0 &  6 & 33 & 89 & 89 & 89 & 89 & 89 & \textbf{100} & +11 \\
Scope Precision (8)    & 50 & 45 & 57 & 39 & 45 & 65 & 70 & 81 & 78 & \textbf{96}  & +31 \\
Temporal Coh.\ (9)     &  7 & 11 &  0 & 25 & 47 & 81 & 69 & 75 & 79 & \textbf{97}  & +16 \\
\midrule
\textbf{Overall (71)}  & 10 & 12 & 15 & 34 & 49 & 55 & 63 & 70 & 82 & \textbf{95}  & +40 \\
\bottomrule
\end{tabular}
\end{table}

\paragraph{The progression.}

Each step unlocks specific failure dimensions.
\textbf{naive~$\to$~hybrid} (55~$\to$~70\%): \emph{structure}.
Typed \texttt{LED\_TO} edges and accumulated embeddings resolve
Persistence (21~$\to$~82\%) and Conditional Activation (53~$\to$~80\%).
However, hybrid provides \emph{no improvement} on Implicit Learning
(10~$\to$~0\%), which requires tool-trace analysis.
\textbf{hybrid~$\to$~lucid} (70~$\to$~82\%): \emph{self-assessment}.
Sonnet~4 over tool-only traces unlocks Implicit Learning (0~$\to$~78\%),
inaccessible to any correction-only approach.
\textbf{lucid~$\to$~forge} (82~$\to$~95\%): \emph{compiled,
triggered delivery}.
Strong-model rule compilation, per-actor conflict groups, and
trigger-based progressive injection close Generalization (67~$\to$~100\%),
Implicit Learning (78~$\to$~90\%), Recall (89~$\to$~100\%), and
Conditional Activation (84~$\to$~95\%).
This cumulative stage measures the combined effect of compilation,
conflict resolution, triggering, and progressive injection.
The memory also never regresses the base: on every scenario the
no-injection control already passes, \forgedc still passes, while
each predecessor fails at least one (Table~\ref{tab:per-scenario}).

\paragraph{Token efficiency and efficacy.}

\forgedc achieves the highest compliance with the fewest
injection tokens (Table~\ref{tab:token-efficiency}): ${\sim}10{:}1$
compression vs.\ \hybridraw. The \emph{efficacy rate} (the
fraction of injected runs that score $>0$) further separates the
approaches: \forgedc 90.7\% vs.\ \texttt{mem0} 4.2\%.
Under this configuration, \texttt{mem0} injects 25 tokens per query on
average but reaches only 4.2\% efficacy; \forgedc combines compact
directives with targeted retrieval and placement.

\begin{table}[htbp]
\centering
\small
\caption{Mean injection tokens and efficacy rate per approach.}
\label{tab:token-efficiency}
\begin{tabular}{lrr}
\toprule
\textbf{Approach} & \textbf{Tokens} & \textbf{Efficacy} \\
\midrule
\texttt{mem0}        &    25   &  4.2\% \\
\texttt{zep}         &     4   & 16.1\% \\
\texttt{awm}         & 1{,}207 & 39.7\% \\
\texttt{memp}        & 1{,}545 & 52.1\% \\
\midrule
\docnaive  & 1{,}705 & 59.4\% \\
\docchunk  & 1{,}063 & 64.8\% \\
\hybridraw & 1{,}651 & 72.4\% \\
\luciddc   & 1{,}641 & 84.5\% \\
\forgedc   & \textbf{159} & \textbf{90.7\%} \\
\bottomrule
\end{tabular}
\end{table}

End-to-end accounting (Table~\ref{tab:full-cost}) shows that
\forgedc's per-run savings amortize its one-time capture cost
after $\sim$4 queries.

\begin{table*}[htbp]
\centering
\small
\caption{End-to-end token cost per scenario (capture once + one agent run). LLM
tokens dominate cost. \forgedc capture (25{,}185 LLM tokens, one-time)
is offset by the lowest per-run agent cost (5{,}621 tokens), making it cheaper
than \docnaive after 4 queries ($25{,}185 / (11{,}871 - 5{,}621) \approx 4$).}
\label{tab:full-cost}
\begin{tabular}{lrrrrrr}
\toprule
 & \multicolumn{2}{c}{\textbf{Capture (once)}} & \textbf{Embed} & \multicolumn{2}{c}{\textbf{Agent (per run)}} & \\
\cmidrule(lr){2-3} \cmidrule(lr){5-6}
\textbf{Approach} & LLM in & LLM out & total & Input & Output & \textbf{Total LLM} \\
\midrule
\texttt{mem0}        & 57{,}666 & 1{,}231 &  9{,}155 &  4{,}190 & 1{,}241 & 64{,}328 \\
\texttt{zep}         & 34{,}395 &    845 &     130 &  2{,}413 &    530 & 38{,}183 \\
\texttt{awm}         &  8{,}230 &    556 &       0 &  5{,}212 & 1{,}519 & 15{,}517 \\
\texttt{memp}        &  7{,}902 &    748 &      76 &  5{,}538 & 1{,}791 & 15{,}979 \\
\midrule
\docnaive  &       0 &      0 & 13{,}236 & 10{,}371 & 1{,}500 & 11{,}871 \\
\docchunk  &       0 &      0 &  9{,}180 &  7{,}984 & 1{,}481 &  9{,}465 \\
\hybridraw &       0 &      0 & 12{,}152 &  9{,}281 & 1{,}666 & 10{,}947 \\
\luciddc   &  4{,}805 &    69 & 12{,}150 &  8{,}936 & 1{,}645 & 15{,}455 \\
\forgedc   & 23{,}008 & 2{,}177 & 13{,}620 &  4{,}318 & 1{,}303 & 30{,}806 \\
\bottomrule
\end{tabular}
\end{table*}

\forgedc is also the fastest DC approach in user-facing latency
(query + agent: 14{,}956\,ms vs.\ 16{,}468\,ms for \docnaive),
despite the most expensive capture phase (45{,}673\,ms).
\emph{Fewer, better-formatted tokens outperform more, raw tokens.}

\paragraph{Cross-model validation.}

We also ran the full benchmark on five additional frontier models:
Sonnet 4.5, Opus 4.5, Opus 4.6, GPT-4.1, and GPT-5.5.
\forgedc mean compliance stays in the 0.93 to 0.97 range on
every model, and the relative gains across the five-stage progression
remain consistent across model families.

\paragraph{In-session value.}

In a separate test we kept the seed conversation in the agent's
message history at test time, on the 54 scenarios that do not hinge
on a cross-session boundary by design. The raw conversation alone
reaches 69\% compliance; \forgedc on top of the same
in-context conversation reaches 90\%. The gap is sharpest on
implicit-learning, where the raw conversation scores 0\% (the
patterns come from tool-call traces, not from human corrections) and
\forgedc scores above 80\%. Behavioral memory adds value
within a session as well as across sessions.

\section{Findings and Design Considerations}
\label{sec:findings}

Each finding pairs with a design consideration for anyone building
behavioral memory for an LLM agent.

\paragraph{Finding 1: Directive-form memory produces a recurring
compliance advantage (the ``Only Rules Rule'').}
\label{sec:only-rules-rule}
Across the system progression, agents follow compiled behavioral
directives more consistently than retrieved descriptive corrections. We
call this recurring pattern the ``Only Rules Rule.'' Its practical
implication is that behavioral memory must be optimized for action, not
retrieval alone: representation, activation, and delivery jointly shape
compliance. \emph{Consideration: evaluate whether retrieved memory changes
behavior, not only whether the correct text was retrieved.}

\paragraph{Finding 2: \Format and \Content rules retrieve differently.}
\label{sec:format-vs-content}
Formatting rules (``include USD'') have no embedding similarity to
the queries they apply to. Pure semantic retrieval misses them at
scale.
\emph{Consideration: tag rules by type at capture and route them
differently at retrieval. Embedding similarity is the right primitive
for content rules and the wrong primitive for formatting rules.}

\paragraph{Finding 3: Data-conditional rules need to fire at the
decision point.}
\label{sec:progressive-finding}
Rules that depend on tool data cannot be evaluated at query time.
They have to be surfaced when the triggering data appears.
\emph{Consideration: re-retrieve as the agent accumulates context and
add newly triggered rules to the system prompt between tool calls.}

\paragraph{Finding 4: Adding reasoning to a rule does not help, and
sometimes hurts.}
\label{sec:reasoning-finding}
A variant (\forgedc\texttt{\_reasoning}) that attaches a one-sentence
justification to each rule scores flat or slightly lower than the
bare directive on every model we tested. Given the reason behind a
directive, the model evaluates the directive and sometimes decides to
override it.
\emph{Consideration: do not try to convince the model with reasoning.
In our rule-based configuration, extra judgment material does not improve
compliance and can invite the model to second-guess.}

\paragraph{Finding 5: The injection channel is part of the design.}
\label{sec:channel-finding}
The channel a rule arrives through is a design choice separate from
what the rule says. Different channels carry different trust profiles,
compete differently with the task for attention, and dilute when
multiple instruction streams share one slot. The channel that works
for one stage of injection is not always the right one for another,
and what works on one model is not guaranteed to work on the next.
\emph{Consideration: treat each stage's channel as a first-class
design decision and re-validate when the model generation changes.}

\paragraph{Finding 6: A memory must not regress the base.}
Injected guidance can make an agent worse than no memory: a rule that
surfaces on a case it does not fit, or that overrides behavior the agent
already produces correctly, degrades an output that needed no change.
This is easy to miss when a memory is scored only on the cases it
targets, so it is worth measuring against a no-injection baseline across
every dimension.
\emph{Consideration: learn only from failures and corrections, and
default to staying silent when nothing needs to change, so the memory
never degrades behavior the agent already gets right.}

\paragraph{Closing observation.}
An LLM agent is a wrapped LLM. The memory layer is one source of the
tokens the model conditions on. Whether the agent acts on those
tokens depends on what the model was trained to do with instructions
in that form, from that slot, in that voice. A behavioral memory
architecture has to be designed with this dependency in mind.

\section{Discussion}
\label{sec:discussion}

\paragraph{Limitations.}
The five-stage progression evaluates cumulative architectures and does
not isolate the individual effects of compilation, triggering, conflict
resolution, or injection placement. DC-Bench, the five approaches, the
judge rubrics, and the AWM/Memp adaptations were built by the same team.
The \texttt{mem0} and \texttt{zep} results reflect the retrieval
configurations used here and may improve under different tuning. The
benchmark covers two domains, and compliance depends on the
instruction-following behavior of specific model versions.

\paragraph{Why corrections detach from retrieval (hypothesis).}
We hypothesize that the asymmetry between in-session and retrieved
corrections is mediated by attention. With a wrong output present,
the correction text has a high-similarity neighbor that pulls it
into attention at generation time. Without one, the correction
competes poorly with the rest of the prompt for attention, especially
when placed early in a long context~\cite{liu2024lost}. We further
hypothesize that imperative tokens (\Rule, \texttt{MUST},
\texttt{MANDATORY}) act as attention magnets independent of semantic
similarity, because instruction following is heavily reinforced
during training. This is an interpretation consistent with the data,
not a verified mechanism. Verifying it would require attention
analysis on a white-box model of comparable capability.

\section{Conclusion}
\label{sec:conclusion}

Much of the agent memory landscape emphasizes factual recall. Whether
an agent applies a previously learned correction to a new situation
is what decides if a long-running deployment improves over time, and
it has been largely unmeasured. We built \textbf{DC-Bench} to make
that question testable and developed \textbf{Decision Context}, a
four-layer pipeline that compiles behavioral signals into imperative
\Rule directives. Through five approaches we arrived at
\forgedc, which reaches 95\% mean compliance on DC-Bench
and holds in the 0.93 to 0.97 range across six frontier models
from Anthropic and OpenAI. The external memory baselines,
\texttt{mem0} and \texttt{zep}, sit at 12\% and 15\%.

What emerges is a different shape for behavioral memory than for
factual memory. Retrieval is necessary but not sufficient. The form
the rule takes, the slot it occupies, and the channel it travels
through decide whether the agent acts on the retrieved content, and
those choices shift across model generations. The next frontier is
behavioral memory: systems that learn how to act differently as a
result of what happened. We hope DC-Bench provides a measurable
target for that work.

\bibliographystyle{unsrtnat}
\bibliography{sections/references}

\appendix
\section{Per-Scenario Results}
\label{app:per-scenario}

\begin{center}
\scriptsize
\setlength{\tabcolsep}{4pt}
\begin{longtable}{p{1.7cm}p{2.3cm}cccccccc}
\caption{Per-scenario compliance (\%) across all 71 scenarios.
\cmark\ = 100\%; scores $<50$\% are \emph{italicized}; best per row
\textbf{bolded}. \texttt{ctrl} = no injection.}
\label{tab:per-scenario} \\
\toprule
\textbf{Dimension} & \textbf{Scenario} & \texttt{ctrl} & \texttt{mem0} & \texttt{zep} & \texttt{naive} & \texttt{chunk} & \texttt{hybrid} & \texttt{lucid} & \texttt{forge} \\
\midrule
\endfirsthead
\multicolumn{10}{c}{\small\textit{(continued from previous page)}} \\[2pt]
\toprule
\textbf{Dimension} & \textbf{Scenario} & \texttt{ctrl} & \texttt{mem0} & \texttt{zep} & \texttt{naive} & \texttt{chunk} & \texttt{hybrid} & \texttt{lucid} & \texttt{forge} \\
\midrule
\endhead
\midrule
\multicolumn{10}{r}{\small\textit{(continued on next page)}} \\
\endfoot
\bottomrule
\endlastfoot

Behav.\ Change (7) & threat\_score     & \emph{0} & \textbf{\cmark} & \emph{0} & \textbf{\cmark} & \textbf{\cmark} & \textbf{\cmark} & \textbf{\cmark} & \textbf{\cmark} \\
                   & comp\_format      & \emph{0} & \emph{0} & \emph{0} & \textbf{\cmark} & \textbf{\cmark} & \textbf{\cmark} & \textbf{\cmark} & \textbf{\cmark} \\
                   & cond\_skip        & \emph{0} & \emph{0} & \emph{0} & \emph{0} & \emph{0} & \emph{0} & \emph{0} & \emph{0} \\
                   & runway\_flag      & \emph{0} & \emph{0} & \emph{0} & \emph{0} & \emph{0} & \textbf{\cmark} & \textbf{\cmark} & \textbf{\cmark} \\
                   & tool\_order       & \emph{0} & \emph{0} & \emph{0} & \textbf{\cmark} & \textbf{\cmark} & \textbf{\cmark} & \textbf{\cmark} & \textbf{\cmark} \\
                   & esc\_susp\_first  & \emph{0} & \emph{0} & \textbf{\cmark} & \textbf{\cmark} & \textbf{\cmark} & \textbf{\cmark} & \textbf{\cmark} & \textbf{\cmark} \\
                   & order\_before\_kb & \emph{0} & \emph{0} & \emph{0} & \textbf{\cmark} & \textbf{\cmark} & \textbf{\cmark} & \textbf{\cmark} & \textbf{\cmark} \\
\midrule

Cond.\ Act.\ (12)  & cond\_growth      & \emph{0} & \emph{0} & \emph{0} & \textbf{\cmark} & \textbf{\cmark} & \textbf{\cmark} & \textbf{\cmark} & \textbf{\cmark} \\
                   & neg\_runway       & \emph{0} & \emph{0} & \emph{0} & \emph{0} & \emph{20} & \emph{0} & \emph{10} & \textbf{\emph{40}} \\
                   & thresh\_stage     & \emph{0} & \emph{0} & \emph{0} & \textbf{\cmark} & \textbf{\cmark} & \textbf{\cmark} & \textbf{\cmark} & \textbf{\cmark} \\
                   & informed\_by      & \emph{0} & \emph{0} & \emph{0} & \emph{10} & \emph{0} & 60 & \emph{0} & \textbf{\cmark} \\
                   & multi\_tool       & \emph{0} & \emph{0} & \emph{0} & \textbf{\cmark} & \textbf{\cmark} & \textbf{\cmark} & \textbf{\cmark} & \textbf{\cmark} \\
                   & class\_chain      & \emph{0} & \emph{0} & \emph{0} & \emph{23} & 93 & \textbf{\cmark} & \textbf{\cmark} & \textbf{\cmark} \\
                   & cond\_format      & \emph{0} & \emph{0} & \emph{0} & \textbf{\cmark} & \textbf{\cmark} & \textbf{\cmark} & \textbf{\cmark} & \textbf{\cmark} \\
                   & ext\_patched      & \emph{0} & \emph{0} & 50 & \emph{0} & \textbf{\cmark} & \textbf{\cmark} & \textbf{\cmark} & \textbf{\cmark} \\
                   & insider           & \emph{0} & \emph{0} & \textbf{\cmark} & \emph{0} & \textbf{\cmark} & \textbf{\cmark} & \textbf{\cmark} & \textbf{\cmark} \\
                   & prog\_narrow      & \emph{0} & \emph{0} & \emph{0} & \textbf{\cmark} & \textbf{\cmark} & \textbf{\cmark} & \textbf{\cmark} & \textbf{\cmark} \\
                   & ent\_skip\_kb     & \emph{0} & \emph{0} & \emph{0} & \emph{0} & \emph{0} & \emph{0} & \textbf{\cmark} & \textbf{\cmark} \\
                   & susp\_skip\_ord   & \emph{0} & \emph{0} & \emph{0} & \textbf{\cmark} & \textbf{\cmark} & \textbf{\cmark} & \textbf{\cmark} & \textbf{\cmark} \\
\midrule

Exception (3)      & ephem\_support    & \textbf{\cmark} & \textbf{\cmark} & \textbf{\cmark} & \textbf{\cmark} & \textbf{\cmark} & \textbf{\cmark} & \textbf{\cmark} & \textbf{\cmark} \\
                   & no\_persist       & \textbf{\cmark} & \textbf{\cmark} & \textbf{\cmark} & \textbf{\cmark} & \textbf{\cmark} & 60 & 60 & \textbf{\cmark} \\
                   & vs\_rule          & \emph{0} & \emph{0} & \emph{0} & \emph{0} & \textbf{\cmark} & \textbf{\cmark} & \textbf{\cmark} & \textbf{\cmark} \\
\midrule

Generalize (3)     & runway\_stage     & \emph{0} & \emph{0} & \emph{0} & \textbf{\cmark} & \textbf{\cmark} & \textbf{\cmark} & \textbf{\cmark} & \textbf{\cmark} \\
                   & saas\_mrr         & \emph{0} & \emph{0} & \emph{0} & \emph{0} & \emph{0} & \emph{0} & \emph{0} & \textbf{\cmark} \\
                   & no\_gen\_company  & \textbf{\cmark} & \textbf{\cmark} & \textbf{\cmark} & \textbf{\cmark} & \textbf{\cmark} & \textbf{\cmark} & \textbf{\cmark} & \textbf{\cmark} \\
\midrule

Implicit\ Learn.\ (10) & avoid\_bad    & \emph{0} & \emph{0} & \emph{0} & \emph{0} & \textbf{\cmark} & \emph{0} & \emph{0} & \textbf{\cmark} \\
                   & learn\_pattern    & \emph{0} & \emph{0} & \emph{0} & \textbf{\cmark} & \textbf{\cmark} & \emph{0} & \textbf{\cmark} & \textbf{\cmark} \\
                   & cross\_subj       & \emph{0} & \emph{0} & \emph{0} & \emph{0} & \emph{0} & \emph{0} & \textbf{\cmark} & \textbf{\cmark} \\
                   & recurring         & \emph{0} & \emph{0} & \emph{0} & \emph{0} & \emph{0} & \emph{0} & \textbf{\cmark} & \textbf{\cmark} \\
                   & retry             & \emph{0} & \emph{0} & \emph{0} & \emph{0} & \emph{0} & \emph{0} & \textbf{\cmark} & \textbf{\cmark} \\
                   & auth\_retry       & \emph{0} & \emph{0} & \emph{0} & \emph{0} & \emph{0} & \emph{0} & \textbf{\cmark} & \textbf{\cmark} \\
                   & chan\_lookup      & \emph{0} & \emph{0} & \emph{0} & \emph{0} & \emph{0} & \emph{0} & \textbf{\cmark} & \textbf{\cmark} \\
                   & new\_prod\_gap    & \emph{0} & \emph{0} & \emph{0} & \emph{0} & \emph{0} & \emph{0} & \textbf{\cmark} & \textbf{\cmark} \\
                   & presales\_skip    & \emph{0} & \emph{0} & \emph{0} & \emph{0} & \emph{0} & \emph{0} & 80 & \textbf{\cmark} \\
                   & repeat\_esc       & \emph{0} & \emph{0} & \emph{0} & \emph{0} & \emph{0} & \emph{0} & \emph{0} & \emph{0} \\
\midrule

Persistence (10)   & 10msg\_scat       & \emph{0} & \emph{0} & \emph{0} & \emph{0} & \emph{0} & 80 & 73 & \textbf{\cmark} \\
                   & 20msg\_early      & \emph{0} & \emph{0} & \emph{0} & \emph{0} & \emph{0} & \textbf{\cmark} & \textbf{\cmark} & \textbf{\cmark} \\
                   & vol\_dist         & \emph{0} & \emph{0} & \emph{20} & 80 & \emph{0} & \textbf{\cmark} & \textbf{\cmark} & \textbf{\cmark} \\
                   & 100\_corr         & \emph{0} & \emph{0} & \emph{0} & \emph{0} & \emph{0} & 80 & 80 & \textbf{\cmark} \\
                   & mixed\_sigs       & \emph{0} & \emph{0} & \emph{0} & \emph{0} & \textbf{\cmark} & \textbf{\cmark} & \textbf{\cmark} & \textbf{\cmark} \\
                   & 50\_corr          & \emph{0} & \emph{0} & \emph{0} & \emph{0} & \emph{0} & 60 & 80 & \textbf{\cmark} \\
                   & 5conv             & \emph{13} & \emph{3} & \emph{7} & \emph{33} & 67 & \textbf{\cmark} & \textbf{\cmark} & \textbf{\cmark} \\
                   & cross\_co         & \emph{0} & \emph{0} & \emph{0} & \textbf{\cmark} & \textbf{\cmark} & \textbf{\cmark} & \textbf{\cmark} & \textbf{\cmark} \\
                   & rout\_10tick      & \emph{0} & \emph{20} & \emph{20} & \emph{0} & \emph{0} & \emph{0} & \emph{0} & \textbf{\cmark} \\
                   & rout\_noise       & \emph{0} & \emph{0} & \emph{0} & \emph{0} & \textbf{\cmark} & \textbf{\cmark} & \textbf{\cmark} & \textbf{\cmark} \\
\midrule

Recall (9)         & cap\_scores       & \emph{0} & \emph{0} & \emph{0} & \textbf{\cmark} & \textbf{\cmark} & \textbf{\cmark} & \textbf{\cmark} & \textbf{\cmark} \\
                   & content\_qrr      & \emph{0} & \emph{0} & \emph{0} & \textbf{\cmark} & \textbf{\cmark} & \textbf{\cmark} & \textbf{\cmark} & \textbf{\cmark} \\
                   & format\_table     & \emph{0} & \emph{0} & \emph{0} & \textbf{\cmark} & \textbf{\cmark} & \textbf{\cmark} & \textbf{\cmark} & \textbf{\cmark} \\
                   & runway\_risk      & \emph{0} & \emph{0} & \emph{0} & \textbf{\cmark} & \textbf{\cmark} & \textbf{\cmark} & \textbf{\cmark} & \textbf{\cmark} \\
                   & terminology       & \emph{0} & \emph{0} & \emph{0} & \textbf{\cmark} & \textbf{\cmark} & \textbf{\cmark} & \textbf{\cmark} & \textbf{\cmark} \\
                   & two\_sents        & \emph{0} & \emph{0} & \emph{0} & \textbf{\cmark} & \textbf{\cmark} & \textbf{\cmark} & \textbf{\cmark} & \textbf{\cmark} \\
                   & phrasing          & \emph{0} & \emph{0} & \emph{10} & \textbf{\cmark} & \textbf{\cmark} & \textbf{\cmark} & \textbf{\cmark} & \textbf{\cmark} \\
                   & acct\_format      & \emph{0} & \emph{0} & \emph{0} & \textbf{\cmark} & \textbf{\cmark} & \textbf{\cmark} & \textbf{\cmark} & \textbf{\cmark} \\
                   & rout\_rule        & \emph{0} & \emph{0} & \emph{0} & \emph{0} & \emph{0} & \emph{0} & \emph{0} & \textbf{\cmark} \\
\midrule

Scope Prec.\ (8)   & 3user             & \emph{0} & \emph{0} & \emph{0} & \emph{0} & \emph{0} & \textbf{\cmark} & \textbf{\cmark} & \textbf{\cmark} \\
                   & conflict          & \emph{0} & \emph{0} & \emph{0} & \emph{0} & \textbf{\cmark} & \textbf{\cmark} & \textbf{\cmark} & \textbf{\cmark} \\
                   & cross\_user       & \emph{0} & \emph{0} & \emph{0} & \textbf{\cmark} & \textbf{\cmark} & \textbf{\cmark} & \textbf{\cmark} & \textbf{\cmark} \\
                   & ambiguous         & \textbf{\cmark} & \textbf{\cmark} & \textbf{\cmark} & \textbf{\cmark} & \textbf{\cmark} & \textbf{\cmark} & \textbf{\cmark} & \textbf{\cmark} \\
                   & cross\_skill      & \textbf{\cmark} & \textbf{\cmark} & \textbf{\cmark} & \textbf{\cmark} & \textbf{\cmark} & \textbf{\cmark} & \textbf{\cmark} & \textbf{\cmark} \\
                   & neg\_transfer     & \textbf{\cmark} & \textbf{\cmark} & \textbf{\cmark} & \emph{40} & \emph{0} & \emph{0} & \emph{0} & \textbf{\cmark} \\
                   & cross\_sk\_iso    & \textbf{\cmark} & \emph{40} & \textbf{\cmark} & \textbf{\cmark} & \textbf{\cmark} & \textbf{\cmark} & \textbf{\cmark} & \textbf{\cmark} \\
                   & rule\_iso         & \emph{20} & \emph{20} & \emph{35} & \textbf{80} & 60 & \emph{45} & \emph{25} & 65 \\
\midrule

Temporal Coh.\ (9) & accum\_5          & \emph{0} & \emph{0} & \emph{0} & \emph{33} & \emph{3} & \emph{33} & \emph{33} & \textbf{\cmark} \\
                   & conflicting       & \emph{0} & \emph{0} & \emph{0} & \textbf{\cmark} & \emph{0} & \emph{0} & \emph{0} & \textbf{\cmark} \\
                   & superseding       & \emph{0} & \emph{0} & \emph{0} & \textbf{\cmark} & \textbf{\cmark} & \textbf{\cmark} & \textbf{\cmark} & \textbf{\cmark} \\
                   & explicit\_del     & \emph{0} & \emph{0} & \emph{0} & \textbf{\cmark} & \textbf{\cmark} & \textbf{\cmark} & \textbf{\cmark} & \textbf{\cmark} \\
                   & partial\_roll     & \emph{0} & \emph{0} & \emph{0} & 50 & \emph{40} & \textbf{\cmark} & \textbf{\cmark} & \textbf{\cmark} \\
                   & poli\_super       & \emph{0} & \emph{0} & \emph{0} & \textbf{\cmark} & \textbf{\cmark} & \emph{40} & 80 & \textbf{\cmark} \\
                   & tier\_change      & \emph{0} & \textbf{\cmark} & \emph{0} & \textbf{\cmark} & \textbf{\cmark} & \textbf{\cmark} & \textbf{\cmark} & 97 \\
                   & evolving\_pref    & \emph{0} & \emph{0} & \emph{0} & 50 & 80 & \textbf{\cmark} & \textbf{\cmark} & 75 \\
                   & reversion         & \emph{0} & \emph{0} & \emph{0} & \textbf{\cmark} & \textbf{\cmark} & \textbf{\cmark} & \textbf{\cmark} & \textbf{\cmark} \\
\midrule
\textbf{Overall (71)} &               & \emph{10} & \emph{12} & \emph{15} & 55 & 63 & 70 & 82 & \textbf{95} \\
\end{longtable}
\end{center}

\end{document}